\pdfoutput=1

\documentclass[11pt,table,xcdraw]{article}
\usepackage{acl} 
\usepackage[utf8]{inputenc}
\usepackage[T1]{fontenc}
\usepackage{microtype}
\usepackage[justification=justified, belowskip=-6pt, font=small]{caption} %
\usepackage[hang,flushmargin]{footmisc}
\usepackage{subfigure}
\usepackage{csquotes}
\usepackage{graphicx}
\usepackage{xspace}
\usepackage{paralist}
\usepackage{times}
\usepackage{amsmath}
\usepackage{amssymb}
\usepackage{pifont}
\usepackage{siunitx}
\usepackage{url}
\usepackage{latexsym}
\usepackage{enumitem}
\usepackage{verbatim}
\usepackage{booktabs}
\usepackage{xcolor}
\usepackage{adjustbox}
\usepackage{algorithm,algpseudocode}
\usepackage{xparse}
\usepackage{multirow}
\usepackage[draft]{todonotes}   
\usepackage{scalefnt}
\usepackage{cleveref}

\makeatletter
\NewDocumentCommand{\LeftComment}{s m}{%
  \Statex \IfBooleanF{#1}{\hspace*{\ALG@thistlm}}\(\triangleright \) #2}
\makeatother

\makeatletter
\def\algbackskip{\hskip-\ALG@thistlm}
\makeatother

\makeatletter
\DeclareRobustCommand\onedot{\futurelet\@let@token\@onedot}
\def\@onedot{\ifx\@let@token.\else.\null\fi\xspace}
\newcommand{\SmallUpperCase}[1]{\textsc{\MakeLowercase{#1}}}   

\makeatletter
\def\hlinewd#1{%
\noalign{\ifnum0=`}\fi\hrule \@height #1 %
\futurelet\reserved@a\@xhline}
\makeatother

\newcommand\smaller[2][0.85]{{\scalefont{#1}#2}}

\makeatletter
\def\hlinewd#1{%
\noalign{\ifnum0=`}\fi\hrule \@height #1 %
\futurelet\reserved@a\@xhline}
\makeatother

\def\ie{\emph{i.e}\onedot} 
 
\def\etc{\emph{etc}\onedot} 
 
\def\sr{\SmallUpperCase{SimCSum}}
\def\srx{\SmallUpperCase{SimCSum \xspace}}

\def\sk{\SmallUpperCase{Spektrum}}
\def\skx{\SmallUpperCase{Spektrum \xspace}}

\def\wk{\SmallUpperCase{Wikipedia}}
\def\wkx{\SmallUpperCase{Wikipedia \xspace}}

\def\ks{\SmallUpperCase{KIS}}
\def\ksx{\SmallUpperCase{KIS \xspace}}

\def\sm{\SmallUpperCase{XLSUM}}
\def\smx{\SmallUpperCase{XLSUM \xspace}}

\def\bg{\SmallUpperCase{BIGBIRD}}
\def\bgx{\SmallUpperCase{BIGBIRD \xspace}}

\def\pg{\SmallUpperCase{PEGASUS}}
\def\pgx{\SmallUpperCase{PEGASUS \xspace}}

\def\lg{\SmallUpperCase{LONG-ED}}
\def\lgx{\SmallUpperCase{LONG-ED \xspace}}

\def\mtv{m\SmallUpperCase{T}\oldstylenums{5}}
\def\mtvx{m\SmallUpperCase{T}\oldstylenums{5} \xspace}

\def\mbr{m\SmallUpperCase{BART}}
\def\mbrx{m\SmallUpperCase{BART \xspace}}

\def\kmb{\SmallUpperCase{KIS}-m\SmallUpperCase{BART}}
\def\kmbx{\SmallUpperCase{KIS}-m\SmallUpperCase{BART \xspace}}

\def\bs{\SmallUpperCase{BS}} 
\def\bsx{\SmallUpperCase{BS \xspace}}

\def\si{\SmallUpperCase{FRE}} 
\def\six{\SmallUpperCase{FRE \xspace}}

\def\rg{\SmallUpperCase{ROUGE}}
\def\rgx{\SmallUpperCase{ROUGE \xspace}}

\def\fs{\SmallUpperCase{F}-score}

\def\ri{\SmallUpperCase{R}\oldstylenums{1}}
\def\rix{\SmallUpperCase{R}\oldstylenums{1} \xspace}

\def\rii{\SmallUpperCase{R}\oldstylenums{2}}
\def\riix{\SmallUpperCase{R}\oldstylenums{2} \xspace}

\def\rl{\SmallUpperCase{RL}}

\title{SimCSum: Joint Learning of Simplification and Cross-lingual Summarization for Cross-lingual Science Journalism}

\author{Mehwish Fatima$^\dagger$, Tim Kolber$^\dagger$, Katja Markert$^\ddagger$ \and Michael Strube$^\dagger$\\
	$^\dagger$Heidelberg Institute for Theoretical Studies, Heidelberg, Germany \\ 
    $^\ddagger$Department of Computational Linguistics, Heidelberg University \\ Heidelberg, Germany \\
	\texttt{(mehwish.fatima|tim.kolber|michael.strube)@h-its.org} \\
    \texttt{markert@cl.uni-heidelberg.de}}
	
\begin{document}
\maketitle
\begin{abstract}
Cross-lingual science journalism generates popular science stories of scientific articles different from the source language for a non-expert audience. Hence, a cross-lingual popular summary must contain the salient content of the input document, and the content should be coherent, comprehensible, and in a local language for the targeted audience. We improve these aspects of cross-lingual summary generation by joint training of two high-level \SmallUpperCase{NLP} tasks, simplification and cross-lingual summarization. The former task reduces linguistic complexity, and the latter focuses on cross-lingual abstractive summarization. We propose a novel multi-task architecture - \srx consisting of one shared encoder and two parallel decoders jointly learning simplification and cross-lingual summarization. We empirically investigate the performance of \srx by comparing it with several strong baselines over several evaluation metrics and by human evaluation. Overall, \srx demonstrates statistically significant improvements over the state-of-the-art on two non-synthetic cross-lingual scientific datasets. Furthermore, we conduct an in-depth investigation into the linguistic properties of generated summaries and an error analysis. 
\end{abstract}

\section{Introduction}\label{sec:intro}

A real-world example of cross-lingual science journalism is Spektrum der Wissenschaft\footnote{\href{https://www.spektrum.de/magazin}{https://www.spektrum.de/magazin}}. It is the German version of Scientific American and an acclaimed bridge between local readers and the latest scientific research in Germany. Spektrum's journalists read English scientific articles and summarize them into popular science stories in German that are comprehensible by local non-expert readers. Spektrum der Wissenschaft approached us to automate the process of their journalist's work. We define cross-lingual science journalism as the fusion of two high-level \SmallUpperCase{NLP} tasks: text simplification and cross-lingual scientific summarization.

Cross-lingual science journalism aims to generate science summaries in a target language from scientific documents in a source language while emphasizing simplification. The readers of science magazines, usually non-experts in scientific fields, can grasp the complex scientific concepts expressed in easy-to-understand language. Moreover, this task is extendable for different readability levels according to age and education: adults, teens and kids, and in various local languages. However, we limit our study to investigate the task for English to German, targeting adult non-expert readers, to automate the process for Spektrum der Wissenschaft.

As no prior work exists to the best of our knowledge, we consider the two closest tasks for finding the recent advancements: monolingual science journalism and cross-lingual summarization\footnote{Cross-lingual scientific summarization is an under-studied area, so we focus on cross-lingual summarization.}. In monolingual science journalism, we discover the trend of taking it a downstream task of abstractive summarization~\citep{dangovski2021,zaman2020} with customized datasets~\citep{zaman2020,goldsack2022}. However, these datasets are not suitable for cross-lingual science journalism. Cross-lingual summarization studies can be divided as pipeline~\citep{ouyang2019,zhu2019,zhu2020} and Multi-Task Learning (\SmallUpperCase{MTL})~\citep{cao2020,bai2021,bai2022} models with synthetic datasets, and direct cross-lingual summarization with non-synthetic datasets~\citep{ladhak2020,fatima2021}. We find that \citet{fatima2021} have collected their datasets for cross-lingual scientific summarization, so we use them to explore the task.

To investigate cross-lingual science journalism, we propose an \SmallUpperCase{MTL}-based model - \srx that jointly trains for simplification and cross-lingual summarization to improve the quality of cross-lingual popular science summaries. \srx consists of one shared encoder and two independent decoders for each task based on a transformer architecture, where we consider cross-lingual summarization as our main task and simplification as our auxiliary task. 
\subsubsection*{Contributions}
We summarize the contributions as follows:  
\begin{enumerate} [topsep=0pt,itemsep=-1ex,partopsep=1ex,parsep=1ex]
    \item We introduce \srx that jointly learns simplification and cross-lingual summarization to improve the quality of cross-lingual science summaries for non-expert readers. We also introduce a strong baseline - Simplify-Then-Summarize to compare the performance of our proposed model.  
    \item We empirically evaluate the performance of \srx against several existing cross-lingual summarization models on two cross-lingual scientific datasets. We also conduct a human evaluation to find the linguistic qualities of generated summaries. 
    \item We further analyze the outputs for various lexical, readability and syntactic-based linguistic features. We also perform error analysis to assess the quality of outputs. 
\end{enumerate}
\section{Related Work}\label{sec:rw}

\subsection{Scientific Summarization}
This section focuses on the datasets for scientific summarization. Most science summarization datasets are collected from English scientific papers paired with abstracts: \SmallUpperCase{ArXiv}~\citep{kim2016,cohan2018}, \SmallUpperCase{PubMed}~\citep{cohan2018,nikolov2018}, \SmallUpperCase{MEDLINE}~\citep{nikolov2018} and science blogs~\citep{vadapalli2018,vadapalli2018sci}. Some work has been conducted for extreme summarization with monolingual dataset~\citep{cachola2020}, extended for cross-lingual extreme summarization~\citep{takeshita2022}. The extreme summarization task generates a one/two-line summary from a scientific abstract/paper, which makes it different from science journalism.

Cross-lingual scientific summarization is an understudied area due to its challenging nature. We find two studies: a synthetic dataset from English to Somali, Swahili, and Tagalog with round trip translation~\citep{ouyang2019}, two real cross-lingual datasets from Wikipedia Science Portal and Spektrum der Wissenschaft for English-German~\citep{fatima2021}. 

\subsection{Cross-lingual Summarization}
This section focuses on \SmallUpperCase{MTL}-based cross-lingual summarization. 
\citet{zhu2019} develop an \SmallUpperCase{MTL} model for English-Chinese cross-lingual summarization. They develop two variations of the transformer model~\citep{vaswani2017}, where the encoder is shared, and two decoders are independent. 
\citet{cao2020} present a \SmallUpperCase{MTL} model for cross-lingual summarization by joint learning of alignment and summarization. Their model consists of two encoders and two decoders, each dedicated to one task while sharing contextual representations. The authors evaluate their model on synthetic cross-lingual datasets for the English-Chinese language pairs.
\citet{takase2020} introduce an \SmallUpperCase{MTL} framework for cross-lingual abstractive summarization by augmenting (monolingual) training data with translations for three pairs: Chinese-English, Arabic-English, and English-Japanese. The model consists of a transformer encoder-decoder model with prompt-based learning in which each training instance is affixed with a special prompt to signal example type. 
\citet{bai2021} develop a variation of multi-lingual \SmallUpperCase{BERT} for English-Chinese cross-lingual abstractive summarization. The model is trained with a few shots of monolingual and cross-lingual examples. 
\citet{bai2022} extend their work by introducing a \SmallUpperCase{MTL} model to improve cross-lingual summaries by combining cross-lingual summarization and translation rates. They add a compression scoring method at the encoder and decoder of their model. They augment their datasets for different compression levels of summaries. One variation consists of cross-lingual and monolingual summarization decoders, while the other consists of cross-lingual and translation decoders. 

Most of these studies focus on English-Chinese synthetic datasets emphasizing summarization and translation. By architecture, \srx is similar to \citet{zhu2019} model as it also consists of one shared encoder and two task-specific decoders.

\subsection{Monolingual Science Journalism}
This section focuses on science journalism models. 
\citet{zaman2020} develop an extension of \SmallUpperCase{PGN}~\citep{see2017} by modifying the loss function, so the model is trained for joint simplification and summarization. It is not a \SmallUpperCase{MTL} model but a summarization model with an added loss for simplification. Moreover, the model is trained on a customized dataset that contains simplified summaries from the Eureka Alert science news website. 

\citet{dangovski2021} introduce science journalism as a downstream task of abstractive summarization and story generation. They apply BERT-based models with a prompting method for data augmentation on a monolingual dataset collected from Science daily press releases and scientific papers. They use three existing models for their work: \SmallUpperCase{Sci-BERT}, \SmallUpperCase{CNN}-based sequence-to-sequence model and story generation model.

We find no similarity between these studies and our work except for the overlap of abstractive summarization, however, we focus on cross-lingual summarization.  
\section{Proposed Model}\label{sec:xssr}
Our model jointly trains for \textbf{Sim}plification and \textbf{C}ross-lingual \textbf{Sum}marization (\sr). 
We first define \SmallUpperCase{MTL} and our tasks, and then discuss the architecture of our proposed model.

\subsection{Multi-Task Learning}
\SmallUpperCase{MTL} is an approach in deep learning which improves generalization by learning different noise patterns from data related to different tasks. We define our \SmallUpperCase{MTL}-based model trained on two tasks: simplification and cross-lingual summarization. We adopt hard parameter sharing as it improves the positive transfer and reduces the risk of overfitting~\citep{ruder2017}.

\subsection{Summarization}
We define single-document abstractive summarization as follows. Given a text \begingroup\makeatletter\def\f@size{10}\check@mathfonts$X\!=\! \left\{x_1,\cdots,x_m\right\}$ \endgroup with $m$ number of sentences comprising of a set of words (vocabulary) \begingroup\makeatletter\def\f@size{10}\check@mathfonts$W_X\!=\!\left\{w_1,\cdots, w_X\right\}$\endgroup, an (encoder-decoder-based) abstractive summarizer generates a summary \begingroup\makeatletter\def\f@size{10}\check@mathfonts$Y\!=\!\left\{y_1,\cdots,y_n\right\}$ \endgroup with $n$ sentences that contain salient information of \begingroup\makeatletter\def\f@size{10}\check@mathfonts$X$\endgroup, where \begingroup\makeatletter\def\f@size{10}\check@mathfonts$m\!\gg\!n$ \endgroup and \begingroup\makeatletter\def\f@size{10}\check@mathfonts$Y$\endgroup consisting of a set of words \begingroup\makeatletter\def\f@size{10}\check@mathfonts$W_{Y}\!=\!\left\{w_1,\cdots, w_{Y}|\,\exists\, w_i \notin W_{X}\right\}$\endgroup. The decoder learns the conditional probability distribution over the given input and all previously generated words, where $t$ denotes the time step.
\begingroup\makeatletter\def\f@size{10}\check@mathfonts
\begin{equation}\label{eq1}
    P_{\theta}(Y|X)=\log P(y_t|\; y_{<t},X)
\end{equation}
\endgroup

Cross-lingual summarization adds another dimension of language for simultaneous translation and summarization. Given a text \begingroup\makeatletter\def\f@size{10}\check@mathfonts$X^{l}\!=\!\left\{x_{1}^{l},\cdots,x_{m}^{l}\right\}$ \endgroup in a language $l$ with $m$ sentences comprising of a vocabulary \begingroup\makeatletter\def\f@size{10}\check@mathfonts$W_{X}^{l}\!=\!\left\{w_{1}^{l},\cdots, w_{X}^{l}\right\}$\endgroup, a cross-lingual summarizer generates a summary \begingroup\makeatletter\def\f@size{10}\check@mathfonts$Y^{k}\!=\!\left\{y_{1}^{k},\cdots, y_{n}^{k}\right\}$ \endgroup in a language $k$ that contains salient information in \begingroup\makeatletter\def\f@size{10}\check@mathfonts$X$\endgroup, where \begingroup\makeatletter\def\f@size{10}\check@mathfonts$m\!\gg\!n$ \endgroup and \begingroup\makeatletter\def\f@size{10}\check@mathfonts$Y$ \endgroup consisting of a vocabulary \begingroup\makeatletter\def\f@size{10}\check@mathfonts$W_{Y}^{k}\!=\!\left\{w_{1}^{k},\cdots, w_{Y}^{k} |\,\exists\, w_i\, \notin W_{X}^{l}\right\}$\endgroup. The conditional probability is the same as in Eq.\ref{eq1}, the only difference being that the language on the decoder side is different from the encoder side.

\begin{figure}[t]
    \centering
    \adjustbox{minipage=\columnwidth}{
    \begin{center}
    \includegraphics[width=\textwidth, keepaspectratio,clip, trim= 0.15cm 0.35cm 0.25cm 0.25cm]{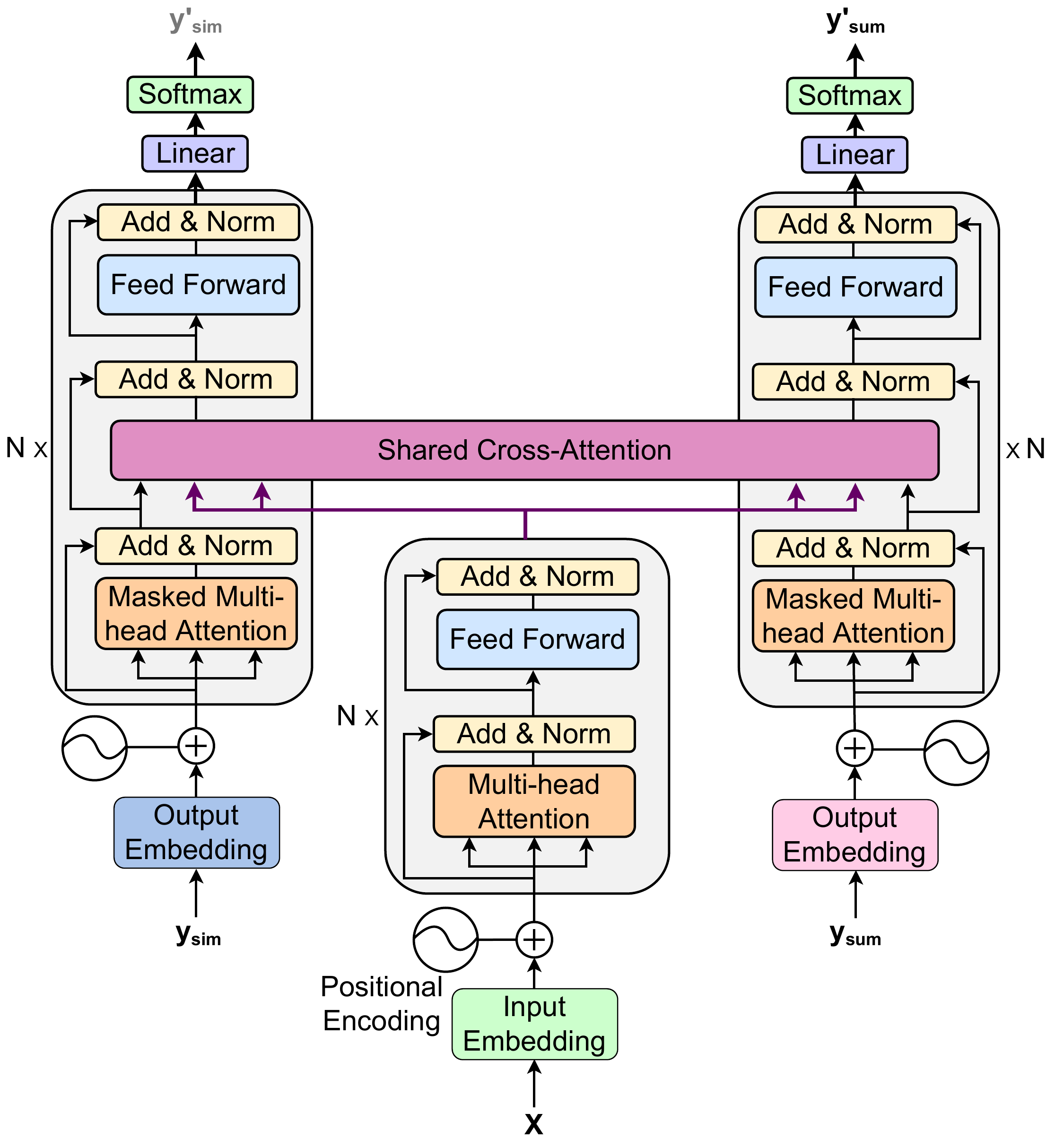}
    \end{center}
    }
    \caption{\srx consists of one shared encoder with two decoding sides for Simplification and cross-lingual Summarization.}\label{fig:ss}
\end{figure}

\subsection{Simplification}
We define the document-level (lexical and syntactic) simplification task as follows. Given a text \begingroup\makeatletter\def\f@size{10}\check@mathfonts$X\!=\!\left\{x_1,\cdots,x_m\right\}$ \endgroup with $m$ sentences comprising of a vocabulary \begingroup\makeatletter\def\f@size{10}\check@mathfonts$W_X\!=\!\left\{w_1,\cdots, w_X\right\}$\endgroup, a simplification model generates the output text \begingroup\makeatletter\def\f@size{10}\check@mathfonts$Y\!=\!\left\{y_1,\cdots, y_n\right\}$ \endgroup that retains the primary meaning of \begingroup\makeatletter\def\f@size{10}\check@mathfonts$X$\endgroup, yet more comprehensible as compared to \begingroup\makeatletter\def\f@size{10}\check@mathfonts$X$\endgroup, where \begingroup\makeatletter\def\f@size{10}\check@mathfonts$m\!\approx\!n$ \endgroup and \begingroup\makeatletter\def\f@size{10}\check@mathfonts$Y$ \endgroup consisting of a vocabulary \begingroup\makeatletter\def\f@size{10}\check@mathfonts$W_{Y}\!=\!\left\{w_1,\cdots, w_{Y} |\, \exists \,w_i \notin W_{X}\right\}$\endgroup. The conditional probability is also the same as in Eq.\ref{eq1}.

\subsection{SimCSum}
We illustrate the framework of \sr\footnote{The code of this work will be public on GitHub upon acceptance.}  in Figure~\ref{fig:ss}. \srx jointly trains on simplification and cross-lingual summarization. \srx adopts hard parameter sharing where the encoder is shared between the tasks while having two task-specific decoders. The decoders only share the cross-attention layer, and the loss is combined to update the parameters ($\theta$). We opt for two decoders because each task's output language and length differ. 
The training method is described in Algorithm~\ref{al}. Here we discuss the further details of \sr. For all mathematical definitions, \begingroup\makeatletter\def\f@size{10}\check@mathfonts$\mathcal{T}\in \left\{sim,sum\right\}$ \endgroup denotes a task.

\begin{algorithm}[t]
\small
  \caption{Training of \srx for Simplification and Cross-lingual Summarization}\label{al}
  \textbf{Input:} 
  \begin{algorithmic}
  \For{\textbf{each} $d \in train set$} 
        \LeftComment{\textit{Process each instance $d$ of dataset $D$ for tuples $I$ of input $x$ and targets for each task $\mathcal{T}$}}
      \State Create $I\langle x,y_{\mathcal{T}}\rangle$ 
    \EndFor
    \\\hrulefill
    \State Initialize model parameters $\theta$
    \State Set maximum Epoch $Ep$
    \For{epoch 1 to $Ep$}
    \For{$b \in train set$}
    \LeftComment{\textit{$b$ is a mini-batch containing I from $train set$}}
    \LeftComment{\textit{\srx consists of Encoder $E$, two Decoders $D_{\mathcal{T}}$}}
    \State Feed $x$ to $E$ and get the cross-attention
    \State Feed $y_{\mathcal{T}}$ to $D_{\mathcal{T}}$
    \State Feed the cross-attention to $D_{\mathcal{T}}$ [\cref{eq:ca}]
    \State $t \leftarrow 0$
    \While {$\theta_{t}$ is not converged}
    \State $t \leftarrow t+1$
    \State Compute $\mathcal{L}(\theta)$ [\cref{eq:loss}]
    \State Compute gradient $\nabla(\theta_{t})$ 
    \State Update $\theta_{t}\leftarrow \theta_{t-1} - \eta\nabla(\theta)$ 
    \EndWhile
    \EndFor
    \EndFor
\end{algorithmic}
\end{algorithm}

\subsubsection{Architecture}
Considering the excellent text generation performance of multi-lingual Bart (\mbr)~\citep{liu2020}, we implement the \srx model based on it and modify it for two decoding sides for each task. Each encoder and decoder stack consists of 12 layers. 

\textbf{Self-Attention.}  
Each layer of encoder/decoder has its self-attention, consisting of keys, values, and queries generated from the same sequence.
\begingroup\makeatletter\def\f@size{10}\check@mathfonts
\begin{align*}
A(Q,K,V) = Softmax(\frac{Q\cdot K^T}{\sqrt{d_k}})\cdot V
\end{align*}
\endgroup
where \smaller{$Q$} is a query, \smaller{$K^T$} is transposed \smaller{$K$} (key) and \smaller{$V$} is the value. All parallel attentions are concatenated to generate multi-head attention scaled with a weight matrix \smaller{$W$}.
\begingroup\makeatletter\def\f@size{10}\check@mathfonts
\begin{align*}
MH(Q,K,V)=Concat(A_1,\cdots, A_h)\cdot W^O
\end{align*}
\endgroup

\textbf{Cross-attention.} 
The cross-attention connects the encoder and decoder and provides the decoder with a weight distribution at each step, indicating the importance of each input token in the current context. We concatenate the cross-attention of both decoders.
\begingroup\makeatletter\def\f@size{10}\check@mathfonts
\begin{align}\label{eq:ca}
A(E,D_{\mathcal{T}}) = Concat(Softmax(\frac{D_{\mathcal{T}}\cdot E^T}{\sqrt{d_k}})\cdot E)
\end{align}
\endgroup
where $E$ is the encoder representation, $D_{\mathcal{T}}$ is the task-specific decoder contextual representation, and $d_k$ is the model size.

\subsubsection{Training Objective}
We train our model end-to-end to maximize the conditional probability of the target sequence given a source sequence. We define the task-specific loss as follows.
\begingroup\makeatletter\def\f@size{10}\check@mathfonts
\begin{align*}
    \mathcal{L}_{\mathcal{T}}(\theta) = \sum_{n=1}^{N}\log P(y_{\mathcal{T}_t}|y_{\mathcal{T}_{<t}}, x; \theta)
\end{align*}
\endgroup
where $x$ represents the input, $y$ is the target, $N$ is the mini-batch size, $t$ is the time step and $\theta$ denotes to learnable parameters. 
We define the total loss of our model by task-specific losses where $\lambda_{\mathcal{T}}$ is an assigned weight to each task.
\begingroup\makeatletter\def\f@size{10}\check@mathfonts
\begin{align}\label{eq:loss}
    \mathcal{L}(\theta) = \sum\lambda_{\mathcal{T}}\cdot\mathcal{L}_{\mathcal{T}}(\theta)
\end{align}
\endgroup
\section{Experiments}\label{sec:exp}
\subsection{Datasets}
We use two non-synthetic cross-lingual scientific summarization datasets. 
\subsubsection{Summarization} 
\textbf{\wk.} 
It is harvested from Wikipedia Science Portal for English and German~\citep{fatima2021}. Wikipedia Science Portal contains articles in various science fields. The \wkx dataset consists of two versions: monolingual and cross-lingual. We use only the cross-lingual part of this dataset. It consists of 50,132 English articles (avg. 1572 words) and German summaries (avg. 100 words). 

\textbf{\sk.} 
It is collected from Spektrum der Wissenschaft~\citep{fatima2021}. Spektrum is a famous science magazine (Scientific American) in Germany. It covers various topics in diversified science fields: astronomy, biology, chemistry, archaeology, mathematics, physics, \etc. The \skx dataset contains 1510 English articles (avg. 2337 words) and German summaries (avg. 361 words).  

\subsubsection{Simplification} 
We construct a synthetic \wkx dataset for the simplification task by applying Keep-It-Simple (\ks)~\citep{laban2021}. To create the simplified \wk, we fine-tune \ksx on \wkx English articles as \ksx is an unsupervised model and does not require parallel data. The simplified \wkx consists of the original English articles paired with simplified English articles.

We perform English text simplification because most of the simplification work has been done in the English language~\citep{al2021}, and very few studies cover the German language~\citep{aumiller2022,weiss2018,hancke2012} for children and dyslexic persons (not suitable for scientific simplification). Moreover, most of the work focuses on lexical or sentence level~\citep{sun2021}. To the best of our knowledge, \ksx is the only \SmallUpperCase{SOTA} paragraph-level unsupervised simplification model. 

\subsection{Split and Usage}
We use \wkx for training, validation and testing (80/10/10), while we use \skx for zero-shot adaptability as a case study.

All \SmallUpperCase{PLM} baselines are trained on \wkx where each instance $I$ in the training set consists of $<\!x,y\!>$ where $x$ is the input English text and $y$ is the target German summary. 

\srx is trained on \wkx where each instance $I$ in the training set contains $<\!x,y_{sim},y_{sum}\!>$ where $x$ denotes the input English article and $y_{sim}$ refers to the simplified English article and $y_{sum}$ is the target German summary.

\subsection{Models} 
\textbf{Baselines.} Almost all cross-lingual \SmallUpperCase{MTL} models in \S\ref{sec:rw} are based on translation and summarization, and none of them applies simplification. So we select several state-of-the-art (\SmallUpperCase{SOTA}) \SmallUpperCase{PLM}s that accept long input texts as baselines. 

We fine-tune the following baselines: (1) \mtv~\citep{xue2021}, (2) \mbr~\citep{liu2020}, (3) \pg~\citep{zhang2020}, (4) LongFormer-Encoder-Decorder (\lg)~\citep{beltagy2020}, and (5) \sm~\citep{hasan2021} and (6) \bg~\citep{zaheer2020}. 

In addition, we define a baseline, Simplify-Then-Summarize, based on \ksx and \mbrx models as a pipeline. We report it as \kmbx in our experiments.

\textbf{SimCSum.} We set \begingroup\makeatletter\def\f@size{10}\check@mathfonts$\lambda_{Sum}\!=\!0.75$ \endgroup for \srx based on the best results on the \wkx validation set.

\subsection{Training and Inference}\label{sec:train}
The libraries, hardware and training time details are presented in Appendix~\ref{app:1}. Here, we discuss hyper-parameters. 

\textbf{Baselines.} We fine-tune all models for a maximum of 25 epochs and average the results of 5 runs for each model. We use a batch size of 4-16, depending on the model size. We use a learning rate (\SmallUpperCase{LR}) of $5e^{-5}$ and 100 warm-up steps to avoid over-fitting of the fine-tuned models. We use the Adam optimizer with a \SmallUpperCase{LR} linearly decayed \SmallUpperCase{LR} scheduler. The encoder language is set to English, and the decoder language is German. 

\textbf{SimCSum.} We adopt similar settings as used for baselines, except for the batch size fixed to 4. 
We only generate tokens from the Summarization decoder side in the inference period. We use beam search of size 5 and a tri-gram block during the decoding stage to avoid repetition.

\subsection{Evaluation}
\textbf{Automatic.} We evaluate all models with three metrics. \rg~\citep{lin2004} is a standard metric for summarization. \SmallUpperCase{Bert}-score~\citep{zhang2019} (\bs) is a recent metric for summarization and simplification as an alternative metric to n-gram-based metrics and applies contextual embeddings. For English text simplification, \SmallUpperCase{SARI} and Flesch Kincaid Grade Level (\SmallUpperCase{FKGL}) are the mostly used metrics~\citep{kariuk2020,omelianchuk2021,laban2021}. As our output language is German, we decide to use a variation of Flesch Kincaid score for the German language, \ie, Flesch Kincaid Reading Ease (\si)~\citep{kincaid1975} (Appendix~\ref{app:2} \S\ref{app:2.2}). 

\textbf{Human.} We conduct a human evaluation to compare the outputs of \srx with \mbrx (baseline) for the same linguistic properties. Our annotators are two university students from the Computational Linguistics department with fluent German and English skills. It is worth mentioning that human evaluation of long cross-lingual scientific text is challenging and costly because it requires bi-lingual annotators with a scientific background. 
\section{Results}\label{sec:results}

\subsection{\wk}
We report \fs\footnote{Precision and Recall of \rgx and \SmallUpperCase{Bert}-score can be found in Table~\ref{tb:a1} in Appendix~\ref{app:6}.} of \rgx and \SmallUpperCase{Bert}-score and \six of all models in Table~\ref{tb:1}. The first block includes the fine-tuned \SmallUpperCase{PLM}s models, the second block presents the pipeline baseline, and the last block includes \sr. 
From Table~\ref{tb:1}, we find that \srx outperforms all baselines for every metric. We compute the statistical significance of the results with the Mann-Whitney two-tailed test for a p-value ($p\!<\!.001$). 
Interestingly, \wkx summaries are not simplified compared to \skx summaries; still, \srx performs better on \wkx than the baselines. We interpret that the simplification auxiliary task helps the \srx to learn better contextual representation and produce more relevant German words. We infer from the results that joint learning of simplification and cross-lingual summarization improves the quality of summaries. 

\begin{table}[t]
    \begin{adjustbox}{width=\columnwidth,center}
\begin{tabular}{llllll}
\toprule
\textbf{\SmallUpperCase{Models}} & \textbf{\ri} & \textbf{\rii} & \textbf{\rl} & \textbf{\bs} & \textbf{\si} \\ 
\midrule
\SmallUpperCase{Gold} & - & - & - & - & 36.93 \\
\mtv & 26.79  & 12.65  & 23.40   & 69.12 & 45.42 \\ 
\mbr & 31.43  & 13.20   & \underline{25.12}  & \underline{70.52} & 44.67 \\ 
\pg & 29.30   & \underline{13.93}  & 24.62  & 69.83 & 43.39 \\ 
\sm & 31.91  & 13.30   & 24.14  & 70.04 & 37.83 \\ 
\bg & 29.23  & 13.72  & 24.60   & 69.19 & 41.42 \\ 
\lg & 15.11  & 06.82   & 13.67  & 63.94 & 24.48 \\ 
\midrule
\kmb  & \underline{32.02}  & 12.39  & 24.72  & \underline{70.52} & \underline{45.76} \\ 
\midrule
\sr & \textbf{34.50$^{\dagger}$ }  & \textbf{14.36$^{\dagger}$} & \textbf{25.85$^{\dagger}$} & \textbf{71.60$^{\dagger}$} & \textbf{46.86$^{\dagger}$} \\ 
\bottomrule
\end{tabular}
\end{adjustbox}
    \caption{The \wkx results for all baselines and \sr. \SmallUpperCase{Gold} denotes the reference summaries. \underline{Underline} refers to the best baseline results, and \textbf{bold}$\dagger$ denotes the best overall results with significant improvements ($p\!<\!.001$).}
    \label{tb:1}
\end{table}

Among the baselines, almost all models demonstrate comparable performance except \lg. For \ri, \ks-\mbrx perform better than other models, however, \mbrx and \smx performance are also similar. \pgx takes the lead for \rii, and \mbrx shows higher performance for \rl. \ks-\mbrx and \mbrx take the lead for \bsx among the baselines. For \si, a score between $30-50$ presents the readability level best understood by college graduates. The \wkx summaries fall in this range. For \si, \ks-\mbrx performs better than the other baselines. Interestingly, almost all baselines except \bgx and \smx demonstrate good performance. Comparing \ks-\mbrx and \mbr, \ks-\mbrx performs slightly better than \mbrx for \rix and \si, equal for \bsx and slightly lower for \riix and \rl. We infer that it is due to the impact of the simplification module in \ks-\mbr. 

\subsection{Case Study: \sk}
Table~\ref{tb:2} presents the results of all models on \sk. 
We find a similar pattern that \srx outperforms all baselines. We also compute the statistical significance of these results with the same procedure. The \skx results are on the lower side compared to the \wkx results due to zero-shot adaptability, especially for \rii. We infer that it is due to the impact of the computation method of \rgx score as it is an n-gram-based metric~\citep{ng2015}. The \skx summaries have higher \six scores compared to \wk. Interestingly, we find that all baselines perform lower than the \SmallUpperCase{Gold} summaries. However, the \srx score is similar to the \SmallUpperCase{Gold} summaries. Comparing the performance of \mbrx and \ks-\mbr, \ks-\mbrx performs slightly lower than \mbrx for all scores except \ri because only \wkx is used for fine-tuning of both models in \ks-\mbr.   

\begin{table}[t!]
    \begin{adjustbox}{width=\columnwidth,center}
\begin{tabular}{llllll}
\toprule
\textbf{\SmallUpperCase{Models}} & \textbf{\ri} & \textbf{\rii} & \textbf{\rl} & \textbf{\bs} & \textbf{\si} \\ 
\midrule
\SmallUpperCase{Gold} & - & - & - & - & 40.76 \\  
\mtv & 09.21   & 00.75   & 06.50   & 58.52 & 38.18 \\ 
\mbr & 16.16  & 01.47   & \underline{13.89}  & 62.11 & 39.17 \\ 
\pg  & 11.49  & 00.95   & 08.01   & 60.56 & 37.93 \\ 
\sm & 17.10   & \underline{01.63}  & 09.79   & \underline{62.25} & 33.83 \\ 
\bg & 12.28  & 01.04   & 08.65   & 59.97 & 36.24 \\ 
\lg & 01.32   & 00.11   & 01.18   & 51.85  & 30.16 \\ 
\midrule
\kmb & \underline{17.33} & 01.61  & 12.97  & 61.83 & \underline{39.21} \\ 
\midrule
\sr   & \textbf{18.88$^{\dagger}$} & \textbf{01.82$^{\dagger}$}  & \textbf{14.16$^{\dagger}$} & \textbf{63.47$^{\dagger}$} & \textbf{40.03$^{\dagger}$} \\ 
\bottomrule
\end{tabular}
\end{adjustbox}

    \caption{The \skx results for all baselines and \sr. \SmallUpperCase{Gold} denotes the reference summaries. \underline{Underline} refers to the best baseline results, and \textbf{bold} with $\dagger$ denotes the best overall results with significant improvements ($p\!<\!.001$).}
    \label{tb:2}
\end{table}

\textbf{Human Evaluation.} We compare the \srx and \mbrx outputs for analyzing linguistic qualities because \sr's architecture is based on \mbr. We provide $30\times 2$ (for each model) random summaries with their original texts. We ask two annotators to evaluate each document for three linguistic properties on a Likert scale from $1-5$. The first five samples are used to calibrate the annotations of annotators, and then each annotator provides independent judgments on the rest of the samples. 

Table~\ref{tb:3h} shows the human evaluation results. The samples used for calibration are not used for computing the scores (guidelines in Appendix~\ref{app:4}). We compute the inter-rater reliability by using Krippendorff's $\alpha$\footnote{\href{https://github.com/LightTag/simpledorff}{https://github.com/LightTag/simpledorff}}. We find that \srx improves the fluency, relevance and readability of outputs. We present a few comparative examples of \srx and \mbrx in Appendix~\ref{app:6}.

\begin{table}[t!]
    \small
    \begin{adjustbox}{width=\columnwidth,center}
\begin{tabular}{llll}
\toprule
\textbf{\SmallUpperCase{Models}} & \textbf{\SmallUpperCase{Fluency}} & \textbf{\SmallUpperCase{Relevance}} & \textbf{\SmallUpperCase{Simplicity}} \\
\midrule
\mbr & 2.28 (0.64) & 1.64 (0.70 ) & 1.86 (0.56) \\
\sr  & 2.62 (0.87) & 2.76 (0.78) & 2.88 (0.81) \\ 
\bottomrule
\end{tabular}
\end{adjustbox}
    \caption{The \skx human evaluation for \mbrx and \sr. The average scores (Krippendorff’s $\alpha$) for each linguistic feature are presented here.}
    \label{tb:3h}
\end{table}
\section{Analysis: \sk}
We explore three further dimensions along with extended readability for in-depth analysis: lexical diversity, syntactic and error types to determine the quality of generated summaries. These types of analysis are well-known in \SmallUpperCase{NLP} for textual analysis~\citep{aluisio2010,hancke2012,vajjala2018,mosquera2022,weiss2022}. The lexical diversity and readability scores are computed over all \sk's reference summaries (Gold) and outputs of \mbrx and \sr. The gold summaries' score is a guideline for how similar the models' outputs are to gold summaries.

\subsection{Lexical Diversity}
Lexical diversity estimates how language is distributed overall and how much cohesion is present in the text as synonyms. It is a good indicator of the readability of a text. We calculate Shannon Entropy Estimation (\SmallUpperCase{SEE})~\citep{shannon1948} and Measure of Textual Lexical Diversity (\SmallUpperCase{MTLD})~\citep{mccarthy2005} to find lexical diversity (see Appendix~\ref{app:2} \S\ref{app:2.1} for formula).

\SmallUpperCase{SEE} presents a text's ``informational value'' and language diversity. It is a language-dependent feature, and its value varies for different languages. Higher \SmallUpperCase{SEE} scores suggest higher lexical diversity. We aim to get similar \SmallUpperCase{SEE} as Gold summaries. Table~\ref{tb:micro} shows \SmallUpperCase{SEE} scores of \mbrx and \srx that are similar to Gold summaries suggesting the similar informational value of all summaries. 

\SmallUpperCase{MTLD} is considered a robust version of the type-token ratio (\SmallUpperCase{TTR}) and calculates lexical diversity with no impact of text length. Higher \SmallUpperCase{MTLD} represents the greater vocabulary richness. Table~\ref{tb:micro} presents \SmallUpperCase{MTLD} scores of \mbrx and \sr. The gold summaries have the highest scores, while \srx is the second highest and \mbrx has the lowest score. These scores suggest that the lexical richness of all three groups is not similar, in contrast to \SmallUpperCase{SEE} results. However, the \srx outputs are more lexically diverse than the \mbrx outputs. We infer from the improved \srx scores that joint learning of simplification and cross-lingual summarization impacts word generation. These results also suggest that \SmallUpperCase{MTLD} provides a better estimation of lexical diversity for our summaries.

\begin{table}[t!]
    \begin{adjustbox}{width=\columnwidth,center}
\begin{tabular}{llll}
\toprule
\textbf{\SmallUpperCase{Features}} & \textbf{\SmallUpperCase{Gold}}  & \textbf{\mbr} & \textbf{\sr}  \\ 
\midrule
\multicolumn{4}{l}{\textbf{Lexical Diversity}} \\ 
\SmallUpperCase{SEE} $\downarrow$ & 4.25 (0.04) & 4.26 (0.1) & 4.25 (0.1)       \\
\SmallUpperCase{MTLD} $\uparrow$ & 201 (41.4) & 65.13 (33.3)  & 91.75 (33.1)     \\ 
\midrule
\multicolumn{4}{l}{\textbf{Readability scores}}     \\ 
\SmallUpperCase{CLI} $\downarrow$ & 18.45 (1.7)   & 21.64 (4.7)   & 20.96 (4.8)      \\
\SmallUpperCase{ARI} $\downarrow$ & 18.99 (2.4)   & 21.07 (5.5)   & 20.26 (5.2)      \\ 
\bottomrule
\end{tabular}
\end{adjustbox}

    \caption{Lexical diversity and readability features' average scores (standard deviation).}
    \label{tb:micro}
\end{table}

\subsection{Readability Scores}
Readability scores measure comprehension levels of the text. One of the syllables-based readability scores is already presented in \S\ref{sec:results}. \citet{coleman1975} suggests that word length in letters is a better predictor of readability than syllables. We calculate Coleman Liau Index (\SmallUpperCase{CLI})~\citep{coleman1975} and Automated Readability Index (\SmallUpperCase{ARI})~\citep{senter1967} as these do not rely on syllables (see Appendix~\ref{app:2} \S\ref{app:2.1} for formula). 

\SmallUpperCase{CLI} computes scores on word lengths. \SmallUpperCase{ARI} computes scores on characters, words and sentences. For both \SmallUpperCase{CLI} and \SmallUpperCase{ARI}, the lower score is better as it shows the ease of reading and understanding. We interpret from Table~\ref{tb:micro} that Gold summaries have the lowest score, \srx has the second-lowest score, and \mbrx has the highest score. We infer from the improved \srx scores that joint learning of simplification and cross-lingual summarization impacts both word and sentence level because \SmallUpperCase{CLI} only focuses on words, while \SmallUpperCase{ARI} includes sentences also. 

\begin{table}[t!]
    \begin{adjustbox}{width=\columnwidth,center}
\begin{tabular}{llll}
\toprule
\textbf{\SmallUpperCase{Features$\downarrow$}} & \textbf{\SmallUpperCase{Gold}} & \textbf{\mbr} & \textbf{\sr}  \\ 
\midrule
ASL  & 24.09 (4.2)  & 24.15 (7.2)   & 20.97 (6.5)     \\
ADD & 3.60 (0.3) & 4.16 (1.1)    & 3.91 (1.1)      \\
ADW  & 0.93 (0.04) & 0.95 (0.02)    & 0.94 (0.04)      \\
ATH & 8.32 (0.7) & 8.72 (1.5)    & 8.57 (1.5)      \\
\bottomrule
\end{tabular}
\end{adjustbox}
    \caption{Syntactic features' average scores (standard deviation).}
    \label{tb:syn}
\end{table}

\subsection{Syntactic Analysis} 
Syntactic analysis elaborates on how words and phrases are related in a sentence structure. We perform it with constituency trees on $25\times 2$ (for each model) random summaries from \mbr, \sr and the gold summaries. The total number of sentences for \mbrx is 70, for \srx is 80 and for gold is 131. Table~\ref{tb:syn} presents four syntactic features (see Appendix~\ref{app:2} \S\ref{app:2.2} for definitions). 

We infer from the average sentence length (\SmallUpperCase{ASL}) that \srx produces shorter sentences than \mbrx and gold summaries, which exhibits syntactic simplicity. A small average dependency distance (\SmallUpperCase{ADD}) shows that words with a dependency relation are close together, making the text easier to understand. Table~\ref{tb:syn} shows that \srx summaries have a smaller average dependency than \mbr, much closer to gold summaries. Fewer dependents per word (\SmallUpperCase{ADW}) make a text less ambiguous and thus easier to follow. Table~\ref{tb:syn} shows the \srx outputs have fewer dependents than the \mbrx outputs and are similar to gold summaries. The average tree height (\SmallUpperCase{ATH}) explains the syntactic structural complexity of a sentence. Table~\ref{tb:syn} shows that \srx outputs are less structurally complex than \mbrx outputs, however, gold summaries have the least average tree height. 
We infer from the syntactic analysis that joint learning of simplification and cross-lingual summarization positively impacts the syntactic properties of summaries. 

\subsection{Error Analysis}
To further explore the challenges of improving cross-lingual science summaries, we randomly select $25\times 2$ (for each model) summaries from the \srx and \mbrx outputs. We find three main categories of errors in the manual inspection. Table~\ref{tb:error} presents the occurrences of these errors in each model. Appendix~\ref{app:5} presents some examples from error analysis and its guidelines.

\textbf{Non-German Words.} It is the error type where the models either produce non-existent German words or partially English-German or another language words. We find that \mbrx is more prone to produce such errors. We infer that it is due to the imbalance between the pre-trained and fine-tuned dataset sizes. These models are pre-trained on many languages and usually fine-tuned on comparatively smaller data. \srx tends to produce fewer errors due to data augmentation (simplification data) during the training. 


\textbf{Wrong Name Entities.} It is the error type where the models produce wrong name entities, such as cities or country names and persons' first and last names. We find that both models tend to produce such errors, however, the percentage of such errors is quite low. We infer that the models overestimate or underestimate the probability of word sequences present in data.

\textbf{Unfaithful Information.} It is the error type where we find some (new) information in generated summaries that is not faithful to the source documents. We infer that this error is caused by long inputs where the model tends to hallucinate and generates some content that cannot be verified from the source. We find that \srx makes similar errors as \mbrx for this error type.

\begin{table}[t!]
    \begin{adjustbox}{width=\columnwidth,center}
\begin{tabular}{lll}
\toprule
\textbf{\SmallUpperCase{Error types}} & \textbf{\mbr} & \textbf{\sr} \\
\midrule
Non-German words & 83 & 35 \\
Wrong name entities & 1 & 2 \\
Unfaithful information & 3 & 3 \\
\bottomrule
\end{tabular}
\end{adjustbox}

    \caption{Error occurrences for \mbrx and \srx summaries which may contain multiple errors.}
    \label{tb:error}
\end{table}

\section{Conclusions}\label{sec:conclusions}
In this paper, we explore the task of cross-lingual science journalism. We propose a novel multi-task model, \sr, that combines two high-level \SmallUpperCase{NLP} tasks, simplification and cross-lingual summarization. \srx jointly trains for reducing linguistic complexity and cross-lingual abstractive summarization. We also introduce a pipeline-based strong baseline for cross-lingual science journalism. Our empirical investigation shows the significantly superior performance of \srx over the \SmallUpperCase{SOTA} baselines on two non-synthetic cross-lingual scientific datasets, also indicated by human evaluation. Furthermore, our in-depth linguistics analysis shows how multi-task learning in \srx has lexical and syntactic impacts on the generated summaries. In the last, we perform error analysis to find what kind of errors has been produced by the model. In the future, we plan to add modules for linguistically informed simplification.
\newpage
\section{Limitations}
We proposed \srx for the Cross-lingual Science Journalism task and verified its performance for \wkx and \skx datasets for the English-German language pair. We believe that \srx is adaptable for other domains and languages. However, we have not verified it experimentally and limited our experiments to English-German scientific texts.

Our model jointly trains on two high-level \SmallUpperCase{NLP} tasks, which takes slightly more time than its base model - \mbr, as it has more parameters to learn during the training. Moreover, our model is trained on synthetic simplification data, which may create a dependency on the simplification model - \ks. Therefore, we plan to add linguistically informed simplification modules in our model in our future work. We also find during error analysis that both \mbrx and \srx have problems (repetition or unfaithful information) with long inputs, which need further investigation.

\section{Ethical Consideration}
\textbf{Reproducibility. } We discussed all relevant parameters, training details, and hardware information in \S\ref{sec:train} and Appendix~\ref{app:1}.

\textbf{Legal Consent.} We obtained legal consent from Spektrum der Wissenschaft to use their dataset. We adopted the public implementations with mostly recommended settings, wherever applicable.


\bibliographystyle{acl_natbib}


\appendix
\newpage
\counterwithin{figure}{section}
\counterwithin{table}{section}

\section{Training and Inference}\label{app:1}

\textbf{Libraries.} 
We train all models with Pytorch\footnote{\href{https://pytorch.org/}{https://pytorch.org/}}, Hugging Face\footnote{\href{https://huggingface.co/}{https://huggingface.co/}} integrated with DeepSpeed\footnote{\href{https://www.microsoft.com/en-us/research/project/deepspeed/}{https://www.microsoft.com/en-us/research/project/deepspeed/}} for parallel model training with ZeRO-2. We apply ZeRO-2\footnote{Initially, we used ZeRO-3 offload with \SmallUpperCase{FP}\oldstylenums{16} evaluation, and the training became quite slow as it consumes a lot of time for offloading during evaluation.} to enable model parallelism. ZeRO-2 reduces the memory footprints for gradients and optimizer because it shards the optimizer states and gradients across \SmallUpperCase{GPU}s. 

\textbf{Hardware.} For all models, we complete training and inference on 4 Tesla \SmallUpperCase{P}\smaller{40} \SmallUpperCase{GPU}s each with \smaller{24}\SmallUpperCase{GB} memory.

\textbf{Training Time.} 
\mbrx takes 1 day and 17 hours, \mtvx takes 10 hours, \pgx takes 1 day and 8 hours, \smx takes 1 day and 3 hours, \lgx takes almost 4 days, and \bgx takes 2 days to complete 25 epochs. \srx takes 2 days to complete 25 epochs. 

\section{Analysis: \sk}\label{app:2}
\subsection{Lexical Diversity}\label{app:2.1}
\SmallUpperCase{SEE} is calculated with a frequency table as follows.
\begingroup\makeatletter\def\f@size{10}\check@mathfonts
\begin{align*}
H(x) = \sum_{i=1}^{n} p(x_i)log_{2}\frac{1}{p(x_i)}
\end{align*}
\endgroup
where $H(x)$ is the total amount of information in an entire probability distribution. $P(x_i)$ refers to the frequency of a token appearing in the text, and $1/p(x)$ denotes the information of each case. 

\SmallUpperCase{MTLD} divides the texts into sequences having the same \SmallUpperCase{TTR} and then calculates the mean length of the sequences. 

\subsection{Readability}\label{app:2.2}
\six is calculated as follows:
\begingroup\makeatletter\def\f@size{10}\check@mathfonts
\begin{align*}
    FRE = 180 - ASL - (58.5 \times ASW)
\end{align*}
\endgroup
where average sentence length (\SmallUpperCase{ASL}) is the number of words divided by the number of sentences in the text. The average number of syllables per word (\SmallUpperCase{ASW}) is the number of syllables divided by the number of words in the text. The numeric values are language-dependent constants.

\SmallUpperCase{CLI} is calculated as follows: 
\begingroup\makeatletter\def\f@size{10}\check@mathfonts
\begin{align*}
    CLI = 5.88\times \frac{L}{W} - 29.6\times \frac{S}{W} - 15.8
\end{align*}
\endgroup
where $L$ is the total number of characters (including numbers and punctuation), $W$ is the total number of words, and $S$ is the total number of sentences in a given text.

\SmallUpperCase{ARI} is computed as follows:
\begingroup\makeatletter\def\f@size{10}\check@mathfonts
\begin{align*}
    ARI =4.71\times \frac{L}{W} + 0.5\times \frac{W}{S} - 21.43
\end{align*}
\endgroup
where $L$ is the total number of characters (including numbers and punctuation), $W$ is the total number of words, and $S$ is the total number of sentences in a given text.

\subsection{Syntactic Analysis}\label{app:2.3}
We use Stanza\footnote{\href{https://stanfordnlp.github.io/stanza/constituency.html}{https://stanfordnlp.github.io/stanza/constituency.html}} to extract dependency relations and Stanford Parser \footnote{\href{https://nlp.stanford.edu/software/lex-parser.shtml}{https://nlp.stanford.edu/software/lex-parser.shtml}} to extract constituency trees for each summary. Before tree generation, we replace all German umlauts (\"a, \"o, \"u and \ss) in the summaries with their replacements (ae, oe, ue and ss) due to encoding issues of the Stanford Parser. 

\textbf{Average Sentence Length.} It is the number of tokens in the sentences averaged over the number of sentences in a summary.

\textbf{Average Dependency Distance.} It is the averaged dependency distance over the sentences, which
means the distance between the dependency heads and their dependents. 

\textbf{Average Dependents per Word.} It computes the average number of dependents for each word. 

\textbf{Average Tree Height.} For computing the average tree height of a summary, we calculate the height of every tree and average it over the sentences.

\section{Human Evaluation}\label{app:4}
\subsection{Task} 
We provided annotators with 30 examples of documents paired with a reference summary and two system-generated summaries. The models' identities had not been revealed. The annotators had to rate each model summary for the following linguistic properties after reading the English document and the German summaries. We asked annotators to use the first 5 examples to resolve the annotator’s conflict and to find a common consensus for rating the linguistic aspects. However, the rest of the examples were annotated independently.  

\subsection{Linguistic Properties}
We asked annotators to annotate each summary for the following linguistic properties. 

\textbf{Relevance.} A summary delivers adequate information about the original text. Relevance determines the content relevancy of the summary.

\textbf{Fluency.} The words and phrases fit together within a sentence, and so do the sentences. Fluency determines the structural and grammatical properties of a summary.

\textbf{Simplicity.} Lexical (word) and syntactic (syntax) simplicity of sentences. A simple summary should have minimal use of complex words/phrases and sentence structure.

\subsection{Scale} 
We use a Likert scale from 1 to 5 to score each property (1:worst | 2:bad | 3:ok | 4:good | 5:best). These scores should be assigned by comparing the outputs of both models. 

\section{Error Analysis}\label{app:5}
\subsection{Guidelines}
We define our informal guidelines for the error analysis as follows. To find the errors in the \mbrx and \srx outputs, we compare them to each other, to the \skx German gold summary and the original English text.

\textbf{Non-German Words.} To find them, it is sufficient to read through our model outputs and look up any unknown words. If one of the unknown words turns out to be a non-German word, we mark them in \textcolor{red}{red}.

\textbf{Wrong name entities.} We find wrong-name entities by comparing the names in both system outputs to the reference summary. If the names differ, we verify with the original text that they refer to the same person and thus represent a mistake by the model, and we mark them in \textcolor{blue}{blue}.

\textbf{Unfaithful information.} We find new/unfaithful information by looking up every piece of information in the model outputs  in the reference summary. We search for this information in the original text, and if it is not present there, it is clear that the model produced new information that is not faithful to the source text. We mark this information in \textcolor{orange}{orange}.

\newpage
\onecolumn
\subsection{Examples}
\begin{table}[h!]
\centering
    \fontsize{10}{11}\selectfont
    \begin{adjustbox}{width=\columnwidth,center}
\begin{tabular}{p{\columnwidth}}
\toprule
\rowcolor[HTML]{D6D9FC} 
\textbf{Target:} im freigehege aufzuwachsen und sich dort im schlamm zu suhlen, stärkt offenbar das immunsystem von schweinen : verglichen mit artgenossen, die in einem stall gehalten wurden, hatten freilandschweine eine höhere anzahl von gesundheitsfördernden darmbakterien. zu diesem ergebnis kommen jetzt wissenschaftler um denise kelly von der university of aberdeen. auch gene, die auf entzündungsreaktionen hindeuten, waren bei tieren, die im außenbereich lebten, wesentlich weniger aktiv. [...] \\ 
\rowcolor[HTML]{D6D9FC} 
\textbf{Translation:} Growing up in an outdoor enclosure and wallowing in the mud there apparently strengthens the immune system of pigs : compared to conspecifics that were kept in a pen, free-range pigs had a higher number of health-promoting intestinal bacteria. scientists led by denise kelly from the university of aberdeen are now coming to this conclusion. genes that indicate inflammatory responses were also significantly less active in animals that lived outdoors. [...] \\ 
\midrule
\rowcolor[HTML]{FEE7E7}
\textbf{\mbr:} als fruehmikrobielle kolonisation wird \textcolor{red}{\textbf{dieabdeckung}} der brustdruese von schweinen bei der fruehen entwicklung des erwachsenen menschen bezeichnet. sie wird in der medizin, der mikrobiologie und der pathologie erforscht. \textcolor{red}{\textbf{early microbial colonization of the gut reduced the incidence of infectious}}, entzuendlichen und \textcolor{red}{\textbf{autoimmunischen diseasen reduced the incidence of infectious, inflammatory and autoimmune diseases.}}  \\
\rowcolor[HTML]{FEE7E7}
\textbf{Translation:} \textcolor{red}{\textbf{the covering}} of the mammary gland of pigs during the early development of the adult human is referred to as early microbial colonization. it is researched in medicine, microbiology and pathology. \textcolor{red}{\textbf{early microbial colonization of the gut reduced the incidence of infectious}}, inflammatory and \textcolor{red}{\textbf{autoimmune diseases reduced the incidence of infectious, inflammatory and autoimmune diseases.}} \\
\midrule
\rowcolor[HTML]{FCDAE8}
\textbf{\srx: } 
die fruehzeitshygiene ist ein begriff aus der entwicklungsbiologie und bezeichnet das phaenomen, dass die fruehzeitliche besiedlung des darmes durch krankheitserreger verhindert wird. die fruehzeitshygiene unterscheidet sich von anderen entwicklungsbiologischen forschungsgebieten wie der entwicklungsphysiologie, der haematologie und der palaeontologie dadurch, dass in ihrer gesamtheit zur fruehen entwicklungsphase die mikrobielle vielfalt des darmes zaehlt.   \\  
\rowcolor[HTML]{FCDAE8}
\textbf{Translation:}
early hygiene is a term from developmental biology and describes the phenomenon that prevents early colonization of the intestines by pathogens. early hygiene differs from other developmental biological research areas such as developmental physiology, haematology and palaeontology in that the microbial diversity of the intestine counts in its entirety for the early development phase. \\
\bottomrule
\end{tabular}
\end{adjustbox}

    \caption{An example of \skx output, where \mbrx produces non-German words (marked as \textcolor{red}{red}) and \srx generates the summary with wrong attention on hygiene. The summaries are translated via Google translate.}
    \label{tab:exp}
\end{table}

\begin{table}[h!]
\centering
    \fontsize{10}{11}\selectfont
    \begin{adjustbox}{width=\columnwidth,center}
\begin{tabular}{p{\columnwidth}}
\toprule
\rowcolor[HTML]{D6D9FC} 
\textbf{Target:} nicht nur stress und genetische faktoren tragen zu essstörungen bei, sondern auch vorübergehende hungerphasen. nun stießen forscher vom europäischen zentrum für hirnforschung in rom auf einen möglichen grund dafür - und zwar in den gehirnen von labormäuse, die sich auch durch schmerzen nicht vom naschen abbringen ließen. die psychologin rossella ventura und ihre kollegen trainierten mäuse darauf, in einer von zwei kammern nach einer belohnung in form von schokolade zu suchen. dann versetzten sie den tieren ausgerechnet in dem raum, wo die leckerei lockte, schmerzhafte elektroschocks. die mäuse mieden fortan die schokolade und hielten sich vorwiegend in der anderen kammer auf - außer jene, die zuvor einige zeit lang kaum futter bekommen hatten. [...] \\ 
\rowcolor[HTML]{D6D9FC} 
\textbf{Translation:} Not only stress and genetic factors contribute to eating disorders, but also periods of starvation. now researchers from the european center for brain research in rome have discovered a possible reason for this - namely in the brains of laboratory mice, which were not deterred from snacking even by pain. psychologist rossella ventura and her colleagues trained mice to look for a chocolate reward in one of two chambers. then they gave the animals painful electric shocks in the room where the treat lured them. From then on, the mice avoided the chocolate and mainly stayed in the other chamber - except for those that had hardly received any food for a while. [...] \\ 
\midrule
\rowcolor[HTML]{FEE7E7}
\textbf{\mbr:}  \textcolor{red}{\textbf{eating disorders}} (engl. ``ess-stoerungen'') nennt man eine vielzahl von ursachen und mechanismen, die zur auspraegung von essstoerungen fuehren koennen.  \\
\rowcolor[HTML]{FEE7E7}
\textbf{Translation:} \textcolor{red}{\textbf{eating disorders}} are a number of causes and mechanisms that can lead to the development of eating disorders. \\
\midrule
\rowcolor[HTML]{FCDAE8}
\textbf{\srx: } 
als essverhalten oder magersucht werden die verhaltensweisen von menschen und tieren bezeichnet, die \textcolor{red}{\textbf{excessive}} (vergoren) oder chronisch (z.b. nach unfaellen oder erkrankungen ) zu es verlangen oder dieses verhalten unterdruecken. essverhalten kann zu verhaltensstoerungen fuehren, die ursache fuer die entstehung von essstoerungen sind innere oder aeussere ursachen des essverhaltens.   \\  
\rowcolor[HTML]{FCDAE8}
\textbf{Translation:} eating habits or anorexia are the behaviors of people and animals that demand \textcolor{red}{\textbf{excessive}} (fermented) or chronic (e.g. after accidents or illnesses) or suppress this behavior. eating behavior can lead to behavioral disorders, the cause of eating disorders are internal or external causes of eating behavior. \\
\bottomrule
\end{tabular}
\end{adjustbox}

    \caption{An example of \skx output, where \mbrx produces non-German words (marked as \textcolor{red}{red}) and quite short summary. While \srx generates also generates a non-German word. The summaries are translated via Google translate.}
    \label{tab:exp}
\end{table}
\begin{table}[h!]
\centering
    \fontsize{10}{11}\selectfont
    \begin{adjustbox}{width=\columnwidth,center}
\begin{tabular}{p{\columnwidth}}
\toprule
\rowcolor[HTML]{D6D9FC} 
\textbf{Target:} der tod kam auf dem höhepunkt der macht : cangrandei. della scala, der berühmteste spross der scaliger aus verona, hatte in zahlreichen kriegerischen auseinandersetzungen die herrschaft über ganz venetien an sich gerissen. doch kurz nach seinem triumphalen einzug in die norditalienische stadt treviso am 18. juli 1329 begannen mit durchfall und erbrechen die ersten symptome einer tödlichen erkrankung.
 [...] \\ 
\rowcolor[HTML]{D6D9FC} 
\textbf{Translation:} death came at the height of power : cangrandei. della scala, the most famous offspring of the scaliger family from verona, seized control of all of veneto in numerous military conflicts. but shortly after his triumphant entry into the northern italian city of treviso on july 18, 1329, the first symptoms of a fatal illness began with diarrhea and vomiting.
 [...] \\ 
\midrule
\rowcolor[HTML]{FEE7E7}
\textbf{\mbr:}  \textcolor{blue}{\textbf{alberto}} cangrande ( * 1291 oder 1329 in verona ; † 29. oktober 1329 ebenda ) war ein italienischer dichter, schriftsteller und politiker der flämischen republik. in der zeit der italienischen republik bestieg er von seinen brudern die dynastie der scaligeri. bei der eroberung der elfenbeinküste 1311 wurde er zum maßgebenden gründer der faction in rom. ab 1314 war er herrscher im königreich venedig. cangrande gilt als vater der italienischen
\\
\rowcolor[HTML]{FEE7E7}
\textbf{Translation:}  \textcolor{blue}{\textbf{alberto}} cangrande ( b. 1291 or 1329 in verona ; † october 29, 1329 ibid ) was an italian poet, writer and politician of the flemish republic. in the time of the italian republic he ascended from his brothers the dynasty of the scaligeri. when he conquered the ivory coast in 1311, he became the main founder of the faction in rome. from 1314 he was ruler of the kingdom of venice. cangrande is considered the father of the italian
\\ 
\midrule
\rowcolor[HTML]{FCDAE8}
\textbf{\srx: } 
cangrande della scala oder cangrande \textcolor{blue}{\textbf{cangrande}} ( * 9. märz 1291 in verona ; † 22. juli 1329 ebenda ) war ein begründer der von den scaligeri beeinflussten dynastie, der unter den herrschern von verona von 1308 bis 1387 regierte. der sohn des veronaer stadtbegründers und stadtmeisters alberto della scala ( genannt „ albert “ oder „ adelsmann “ ), der nur als unterstellt war, wurde nach seinem tod 1301 zum bruder von alberto verbannte.  \\  
\rowcolor[HTML]{FCDAE8}
\textbf{Translation:} cangrande della scala or cangrande \textcolor{blue}{\textbf{cangrande}} ( born march 9, 1291 in verona ; † july 22, 1329 ibid ) was a founder of the dynasty influenced by the scaligeri, who reigned under the rulers of verona from 1308 to 1387. the son of verona's founder and mayor alberto della scala ( called " albert " or " nobleman " ), who was only subordinate, was banished to alberto's brother after his death in 1301.\\
\bottomrule
\end{tabular}
\end{adjustbox}

    \caption{An example of \skx output, where \mbrx generates a wrong named entity. \sk, on the other hand, gets it right but generates a wrong alias for this person. The summaries are translated via Google translate.}
    \label{tab:exp}
\end{table}
\begin{table}[h!]
\centering
    \fontsize{10}{11}\selectfont
    \begin{adjustbox}{width=\columnwidth,center}
\begin{tabular}{p{\columnwidth}}
\toprule
\rowcolor[HTML]{D6D9FC} 
\textbf{Target:} inoffizieller weltrekordhalter im nicht - frieren trotz eisiger kälte ist der niederländer wimhof : er lief bereits ziemlich nackt halbmarathons nördlich des polarkreises und nahm das längste eisbad seit beginn der aufzeichnungen - ohne sich ernsthaft zu beklagen. wissenschaftliche untersuchungen des eismannes hatten bisher keine körperlichen auffälligkeiten gefunden, die erklären könnten, warum hof mit kälte offensichtlich besser umgeht als der durchschnittsmensch. [...] \\ 
\rowcolor[HTML]{D6D9FC} 
\textbf{Translation:} the unofficial world record holder for not freezing despite the icy cold is the dutchman wimhof : he ran half marathons north of the arctic circle quite naked and took the longest ice bath since records began - without seriously complaining. scientific examinations of the ice cream man had not found any physical abnormalities that could explain why hof obviously deals with the cold better than the average person.
 [...] \\ 
\midrule
\rowcolor[HTML]{FEE7E7}
\textbf{\mbr:}  wim hof ( * 20. april 1959 ) ist ein niederländischer sportler, der sich unter anderem \textcolor{orange}{\textbf{mit eisschwimmern beschäftigt}}.\\
\rowcolor[HTML]{FEE7E7}
\textbf{Translation:}  wim hof ( born april 20, 1959 ) is a dutch athlete who is \textcolor{orange}{\textbf{involved with ice swimmers}}, among other things.
\\ 
\midrule
\rowcolor[HTML]{FCDAE8}
\textbf{\srx: } 
wim hof ( * 20. april 1959 ), auch als eismann bekannt, ist ein niederländischer \textcolor{orange}{\textbf{swimmer und ein ehemaliger olympiateilnehmer}}. er hält seit dem \textcolor{orange}{\textbf{28. februar 2015}} den bislang längsten direkten kontakt menschlichen körpers mit eis.  \\  
\rowcolor[HTML]{FCDAE8}
\textbf{Translation:} Wim Hof ( born April 20, 1959 ), also known as Eismann, is a Dutch \textcolor{orange}{\textbf{swimmer and a former Olympic competitor}}. \textcolor{orange}{\textbf{since february 28, 2015}}, he has been in the longest direct contact between the human body and ice. \\
\bottomrule
\end{tabular}
\end{adjustbox}

    \caption{An example of \skx output, where both \mbrx and \srx produce unfaithful information. Marked in \textcolor{orange}{orange} is unfaithful information to the original text. The summaries are translated via Google translate.}
    \label{tab:exp}
\end{table}


\newpage
\section{\srx Examples and Results}\label{app:6}
Here we present some examples showing the difference between \srx and \mbr.

\begin{table}[h!]
\centering
    \fontsize{10}{11}\selectfont
    \begin{adjustbox}{width=\columnwidth,center}
\begin{tabular}{p{\columnwidth}}
\toprule
\rowcolor[HTML]{D6D9FC} 
\textbf{Target:} für menschen ist der einbeinige stand immer eine wackelpartie , vor allem bei geschlossenen augen . um nicht umzukippen , müssen die muskeln permanent die leichten schwankungen ausgleichen . nicht so bei flamingos : sie kostet es weniger kraft , auf einem bein zu stehen als auf zweien . deswegen können sie auch beruhigt auf einem bein ein schläfchen machen , ohne dabei umzukippen . [...] \\ 
\rowcolor[HTML]{D6D9FC} 
\textbf{Translation:} standing on one leg is always a shaky game for humans, especially with closed eyes. in order not to tip over , the muscles have to constantly compensate for the slight fluctuations . Not so with flamingos: it takes less strength for them to stand on one leg than on two. that's why you can take a nap on one leg without tipping over.  [...] \\ 

\midrule
\rowcolor[HTML]{FEE7E7}
\textbf{\mbr:}  
\textcolor{orange}{\textbf{die biogerontologie ( von griech. bíos „ leben “ und lat. protes „ befestigt “ ) beschäftigt sich mit dem verhältnis von körpergewicht und körperhaltung. so untersucht die biogerontologie das verhältnis von körpergewicht und körperhaltung : welche gewichtszustände und welche muskeln notwendig sind, um einen pilz, der einer krankheit ausgesetzt ist, zu tragen? wie sehr wichtig es ist, den jeweiligen körpergewichtszustand zu messen. dies wird in der}} 
\\
\rowcolor[HTML]{FEE7E7}
\textbf{Translation:}  \textcolor{orange}{\textbf{biogerontology (from greek bíos “life” and lat. protes “fixed”) deals with the relationship between body weight and posture. this is how biogerontology examines the relationship between body weight and posture: which weight conditions and which muscles are necessary to carry a fungus that is exposed to a disease? how very important it is to measure the respective body weight condition. this will be in the}} 
\\ 
\midrule
\rowcolor[HTML]{FCDAE8}
\textbf{\srx: } 
flamingos ( phoenicopteridae ) oft sitzen auf einem bein, um eine muskelkontraktion zu erzeugen. sie haben die fähigkeit, das körpergewicht aufzunehmen und zu tragen, ohne dabei eine muskelaktive wirkung auszuüben.
\\  
\rowcolor[HTML]{FCDAE8}
\textbf{Translation:} Flamingos (phoenicopteridae) often perch on one leg to produce muscle contraction. they have the ability to absorb and carry body weight without exerting a muscle-active effect. \\
\bottomrule
\end{tabular}
\end{adjustbox}

    \caption{An example of \skx output, where \srx generates a better summary than \mbrx. In this case,  the \mbrx summary misses the article's point by focusing on biogerontology which has no relevancy in this case, and \srx generates a summary that better captures the idea and main focus of the original text. Marked in \textcolor{orange}{orange} is unfaithful information to the original text. The summaries are translated via Google translate.}
    \label{tab:exp}
\end{table}
\begin{table}[h!]
\centering
    \fontsize{10}{11}\selectfont
    \begin{adjustbox}{width=\columnwidth,center}
\begin{tabular}{p{\columnwidth}}
\toprule
\rowcolor[HTML]{D6D9FC} 
\textbf{Target:} eine besondere strategie, seinen fressfeinden zu entkommen, hat der bombardierkäfer pheropsophus jessoensis entwickelt. er wehrt sich sogar noch, nachdem er eigentlich schon verloren hat, und das sogar recht erfolgreich, wie jetzt shinji sugiura und takuya sato von der universität kobe in den biology letters berichten. gerade verschluckt, schießt der käfer im magen der kröte einen strahl heißes, ätzendes sekret aus der afterdrüse - eine strategie, die den bombardierkäfern ihren namen eingebracht hat. normalerweise allerdings dient die waffe zur abschreckung. in diesem fall jedoch ätzen sich die käfer mit ihrer hilfe quasi zurück in die freiheit : das sekret lässt die kröten erbrechen ; und der eigentlich schon verspeiste käfer entkommt. [...] \\ 
\rowcolor[HTML]{D6D9FC} 
\textbf{Translation:} the bombardier beetle pheropsophus jessoensis has developed a special strategy for escaping its predators. he even defends himself after he has actually already lost, and even quite successfully, as shinji sugiura and takuya sato from the university of kobe now report in the biology letters. just swallowed, the beetle in the toad's stomach shoots a jet of hot, caustic secretion out of the anal gland - a strategy that has given the bombardier beetle its name. normally, however, the weapon serves as a deterrent. in this case, however, the beetles etch their way back to freedom with your help: the toads vomit the secretion; and the actually already eaten beetle escapes.
 [...] \\ 
\midrule
\rowcolor[HTML]{FEE7E7}
\textbf{\mbr:}  
\textcolor{orange}{\textbf{unter zoologischer kompatibilitaet versteht man in der biologie die faehigkeit von beutetieremischlingen, nach einem laubabschweif von artgenossen in den verdauungstrakt von artgenossen zu entweichen. die oekologische kompatibilitaet}}  ist ein wesentlicher bestandteil der erfolgreichen entstehung von beutegreifern.beim laubabschweifexperiment mit hilfe des insektenpheropsophus jessoensis aus zwei kaeferarten, dem bufo japonicus und dem b. torrenticola, konnte nachgewiesen werden, dass der kaefer pheropsophus jessoensis unter laboratorischen bedingungen eine in den magensaftgastropfen der jagdsekten einschleudern kann.
\\
\rowcolor[HTML]{FEE7E7}
\textbf{Translation:}  in biology, \textcolor{orange}{\textbf{zoological compatibility means the ability of hybrid prey animals to escape into the digestive tract of conspecifics after the leaves have drifted away from conspecifics. ecological compatibility}}  is an essential part of the successful emergence of predators.in the leaf-dropping experiment with the help of the insect pheropsophus jessoensis from two beetle species, the bufo japonicus and the b. torrenticola, it could be proven that the beetle pheropsophus jessoensis can inject one into the gastric juice droplets of the hunting sects under laboratory conditions.
\\ 
\midrule
\rowcolor[HTML]{FCDAE8}
\textbf{\srx: } 
als wehrverhalten oderverteidigungsverhalten bezeichnet man in der verhaltensbiologie verschiedene verhaltensweisen von tieren, die wehren gegen die fressfeinde oder andere fresstiere bilden. nicht zu verwechseln ist das wehrverhalten 
\textcolor{orange}{\textbf{mit der phylogenetik, die sich aus der evolutionsgeschichte entwickelt hat.}}
\\  
\rowcolor[HTML]{FCDAE8}
\textbf{Translation:} in behavioral biology, defensive behavior is the term used to describe various behaviors of animals that defend themselves against predators or other predators. the defensive behavior should not be confused with \textcolor{orange}{\textbf{the phylogenetics that have developed from the history of evolution.}} \\
\bottomrule
\end{tabular}
\end{adjustbox}

    \caption{An example of \skx output, both \srx and \mbrx produce reasonable summaries. The \srx summary is preferred because it explains the basic idea of the beetle defending itself against predators. The \mbrx summary is filled with complex words that do not make sense or are unnecessary for this context. These words are marked in \textcolor{orange}{orange}. The summaries are translated via Google translate.}
    \label{tab:exp}
\end{table}

\begin{table}[h!]
\centering
    \fontsize{10}{11}\selectfont
    \begin{adjustbox}{width=\columnwidth,center}
\begin{tabular}{p{\columnwidth}}
\toprule
\rowcolor[HTML]{D6D9FC} 
\textbf{Target:} manche farben bleiben besser im gedächtnis haften als andere, doch auf die erinnerung an den entsprechend gefärbten gegenstand hat das keinen einfluss. das zeigt nun eine studie von psychologen um christof kuhbandner von der universität regensburg. die ergebnisse machen deutlich, dass der gegenstand selbst und dessen eigenschaften vom gehirn nicht immer als einheit betrachtet werden. die wissenschaftler testeten für ihre studie das gedächtnis ihrer versuchspersonen. dabei variierten sie sowohl, was zu merken war, als auch, wie es dargeboten wurde. insbesondere präsentierten sie wörter und gegenstände in unterschiedlichen farben. [...] \\ 
\rowcolor[HTML]{D6D9FC} 
\textbf{Translation:} some colors stick in the memory better than others, but this has no influence on the memory of the correspondingly colored object. this is now shown by a study by psychologists led by christof kuhbandner from the university of regensburg. the results make it clear that the object itself and its properties are not always viewed as a unit by the brain. for their study, the scientists tested the memory of their subjects. they varied both what was to be remembered and how it was presented. in particular, they presented words and objects in different colors.
 [...] \\ 
\midrule
\rowcolor[HTML]{FEE7E7}
\textbf{\mbr:}  
die farbpsychologie ist ein teilgebiet der psychologie und beschäftigt sich mit der wirkung von farben auf die wahrnehmung und das gedächtnis.
\\
\rowcolor[HTML]{FEE7E7}
\textbf{Translation:}  color psychology is a branch of psychology and deals with the effect of colors on perception and memory.
\\ 
\midrule
\rowcolor[HTML]{FCDAE8}
\textbf{\srx: } 
die farbe ist ein farbmuster, das uns wahrnehmungen und die sinneswahrnehmung durch das sehen vermittelt. die farbwahrnehmung gehört zu den kognitionswissenschaften. \textcolor{orange}{\textbf{wir sehen sie nicht nur als eindimensionales spektrum möglicher farben, sondern auch als eindimensionales, oftmals auch als „ visuelles system “, das die sinneswahrnehmung mitanpasst. }} [...]  \\ 
\\  
\rowcolor[HTML]{FCDAE8}
\textbf{Translation:} color is a color pattern that conveys perceptions and sensory perception to us through seeing. Color perception is one of the cognitive sciences. \textcolor{orange}{\textbf{we see them not only as a one-dimensional spectrum of possible colors, but also as a one-dimensional, often also as a " visual system " that also adapts the sensory perception.}} [...]  \\ 
\bottomrule
\end{tabular}
\end{adjustbox}
    \caption{An example of \skx output, where \mbrx performs better than \srx. \mbrx generates a summary that is too short but which better recapitulates the main idea. The \textcolor{orange}{orange} marked words in the \srx summary are incoherent and are not faithful to the original text. The summaries are translated via Google translate.}
    \label{tab:exp}
\end{table}
\newpage

\begin{table}[h!]
    \small
    \begin{adjustbox}{width=\columnwidth,center}
\begin{tabular}{l|lll|lll|lll|lll|l}
\toprule
\textbf{\SmallUpperCase{Models}} & \multicolumn{3}{c|}{\textbf{\ri}} & \multicolumn{3}{c|}{\textbf{\rii}} & \multicolumn{3}{c|}{\textbf{\rl}} & \multicolumn{3}{c|}{\textbf{\bs}} & \textbf{\si} \\ 
  & F & P & R & F & P & R & F & P & R & F & P & R & \\ 
\midrule
\multicolumn{14}{c}{\textbf{\SmallUpperCase{Wikipedia}}} \\
\midrule
\mtv & 26.79 & 38.64 & 20.50 & 12.65 & 17.23 & 9.99 & 23.4 & 33.65 & 17.94 & 69.12 & 72.32 & 66.20 & 45.42 \\ 
\mbr & 31.43 & 36.40 & 27.66 & 13.20 & 15.22 & 11.65 & \underline{25.12} & 28.08 & 22.73 & \underline{70.52} & 71.22 & 69.83 & 44.67 \\ 
\pg & 29.30 & 43.01 & 22.22 & \underline{13.93} & 19.81 & 10.74 & 24.62 & 35.82 & 18.76 & 69.83 & 73.32 & 66.66 & 43.39 \\ 
\sm & 31.91 & 36.93 & 28.09 & 13.30 & 16.83 & 11.00 & 24.14 & 29.58 & 20.39 & 70.04 & 71.45 & 68.68 & 37.83 \\ 
\bg & 29.23 & 40.92 & 22.74 & 13.72 & 19.13 & 10.69 & 24.6 & 34.28 & 19.18 & 69.19 & 72.26 & 66.37 & 41.42 \\ 
\lg & 15.11 & 46.43 & 9.02 & 6.82 & 19.57 & 4.13 & 13.67 & 41.33 & 8.19 & 63.94 & 70.87 & 58.25 & 24.48 \\ 
\midrule
\kmb & \underline{32.02} & 33.66 & 30.54 & 12.39 & 13.22 & 11.65 & 24.72 & 27.08 & 22.73 & \underline{70.52} & 71.22 & 69.83 & \underline{45.76} \\ 
\midrule
\sr & \textbf{34.50$^{\dagger}$} & 36.17 & 32.97 & \textbf{14.36$^{\dagger}$} & 15.13 & 13.66 & \textbf{25.85$^{\dagger}$} & 27.07 & 24.73 & \textbf{71.60$^{\dagger}$} & 72.35 & 70.87 & \textbf{46.86$^{\dagger}$} \\ 
\midrule
\multicolumn{14}{c}{\textbf{\SmallUpperCase{Spektrum}}} \\
\midrule
\mtv & 9.21 & 22.15 & 5.81 & 0.75 & 1.89 & 0.47 & 6.5 & 15.84 & 4.09 & 58.52 & 61.84 & 55.53 & 39.18 \\ 
\mbr & 16.16 & 28.23 & 11.32 & 1.47 & 2.66 & 1.02 & \underline{13.89} & 20.06 & 10.62 & 62.11 & 64.66 & 59.75 & 39.17 \\ 
\pg & 11.49 & 25.77 & 7.39 & 0.95 & 2.14 & 0.61 & 8.01 & 18.27 & 5.13 & 60.56 & 63.7 & 57.72 & 37.93 \\ 
\sm & 17.10 & 26.98 & 12.52 & \underline{1.63} & 2.68 & 1.17 & 9.79 & 16.92 & 6.89 & \underline{62.25} & 64.52 & 60.13 & 33.83 \\ 
\bg & 12.28 & 22.49 & 8.45 & 1.04 & 2.07 & 0.69 & 8.65 & 15.98 & 5.93 & 59.97 & 62.46 & 57.68 & 36.24 \\ 
\lg & 1.32 & 1.18 & 1.49 & 0.11 & 0.09 & 0.13 & 1.18 & 1.07 & 1.32 & 51.85 & 50.49 & 53.29 & 30.16 \\ 
\midrule
\kmb & \underline{17.33} & 38.55 & 11.18 & 1.61 & 4.29 & 0.99 & 12.97 & 22.22 & 9.16 & 61.83 & 65.41 & 58.63 & \underline{39.21} \\ 
\midrule
\sr & \textbf{18.88$^{\dagger}$} & 26.03 & 14.81 & \textbf{1.82$^{\dagger}$} & 2.56 & 1.41 & \textbf{14.16$^{\dagger}$} & 16.91 & 12.18 & \textbf{63.47$^{\dagger}$} & 65.22 & 61.81 & \textbf{40.03$^{\dagger}$} \\ 
\bottomrule
\end{tabular}
\end{adjustbox}

    \caption{The \wkx and \skx results for all baselines and \sr. \SmallUpperCase{Gold} denotes the reference summaries. \underline{Underline} refers to the best baseline results, and \textbf{bold}$\dagger$ denotes the best overall results with significant improvements ($p\!<\!.001$).}
    \label{tb:a1}
\end{table}

\end{document}